\documentclass[letterpaper]{article} % DO NOT CHANGE THIS
\usepackage{aaai25}  % DO NOT CHANGE THIS
\usepackage{times}  % DO NOT CHANGE THIS
\usepackage{helvet}  % DO NOT CHANGE THIS
\usepackage{courier}  % DO NOT CHANGE THIS
\usepackage[hyphens]{url}  % DO NOT CHANGE THIS
\usepackage{graphicx} % DO NOT CHANGE THIS
\usepackage{natbib}  % DO NOT CHANGE THIS AND DO NOT ADD ANY OPTIONS TO IT
\usepackage{caption} % DO NOT CHANGE THIS AND DO NOT ADD ANY OPTIONS TO IT
\usepackage{booktabs}

\usepackage{algorithm}
\usepackage{algorithmic}

\usepackage{newfloat}
\usepackage{listings}
\DeclareCaptionStyle{ruled}{labelfont=normalfont,labelsep=colon,strut=off} % DO NOT CHANGE THIS
\floatstyle{ruled}
\newfloat{listing}{tb}{lst}{}
\floatname{listing}{Listing}
\title{Enhancing Document Key Information Localization Through Data Augmentation}
\author{
    Yue Dai
}
\affiliations{
    The University of Western Australia\\
    35 Stirling Hwy, Crawley WA 6009, Australia\\
    yue.dai@research.uwa.edu.au
}

\usepackage{bibentry}
\begin{document}

\maketitle

\begin{abstract}
     The Visually Rich Form Document Intelligence and Understanding (VRDIU) Track B\footnote{\url{https://www.kaggle.com/competitions/vrd-iu2024-trackb}} focuses on the localization of key information in document images. The goal is to develop a method capable of localizing objects in both digital and handwritten documents, using only digital documents for training. This paper presents a simple yet effective approach that includes a document augmentation phase and an object detection phase. Specifically, we augment the training set of digital documents by mimicking the appearance of handwritten documents. Our experiments demonstrate that this pipeline enhances the models' generalization ability and achieves high performance in the competition.
\end{abstract}

\section{Introduction}
As document digitization becomes increasingly prevalent for conveying information, automatically understanding documents has become a crucial task in daily life. Various tasks and datasets have been developed for document understanding, such as document question answering \cite{mathew2021docvqa,ding2024mvqa}, information extraction \cite{wang2021towards,ding2024m3}, and information localization \cite{vsimsa2023docile}. Among these, document information localization requires the model to comprehend document layout to accurately locate the required entities. However, the diversity in document styles—such as digital, printed, and handwritten—introduces complexity to the task due to varying domains. To mitigate the domain gaps, we designed a pipeline that starts with clean digital documents and enriches the dataset through document-oriented data augmentation. This approach has improved the generalizability of adopted frameworks in information localization for handwritten documents, even when the model is trained only on digital ones.
\section{Related Work}
\textbf{Image augmentation} has been a prevalent technique in computer vision tasks, with traditional methods including rotation, cropping, color space transformation, Gaussian noise injection, and more \cite{shorten2019survey}. However, these techniques are not specifically designed for document images, nor can they effectively mimic different document styles. Augraphy, introduced by \cite{groleau2023augraphy}, is a Python library that addresses this gap by offering 26 unique augmentations specifically designed for document images.

\textbf{Layout Analysing} plays a vital role in understanding the structure of documents, as it involves detecting and recognizing semantic entities to transform unstructured documents into structured formats. Initially, vision-only frameworks like Faster R-CNN \cite{ren2015faster} and Mask R-CNN \cite{he2017mask} were commonly used as foundational models for this task. However, to gain a deeper and more nuanced understanding, multi-modal \cite{huang2022layoutlmv3} and relation-aware \cite{luo2022doc} frameworks have been developed, leading to significant improvements. These advanced models can enhance existing object detection frameworks by offering richer and more comprehensive representations, making them powerful backbones for document layout analysis.

\section{Methodology}
\subsection{Dataset}
The pipeline is tested on the public dataset Form-NLU \cite{ding2023form}. We follow the setup from the VRDIU Track B\cite{vrd-iu2024-trackb}, where the training and validation sets consist only of digital documents, while the test set includes both digital and handwritten documents. This setup presents the unique challenge of dealing with domain differences between the training and test data.

\subsection{Data Preprocessing}
\begin{figure*}
    \centering
    \includegraphics[width=\linewidth]{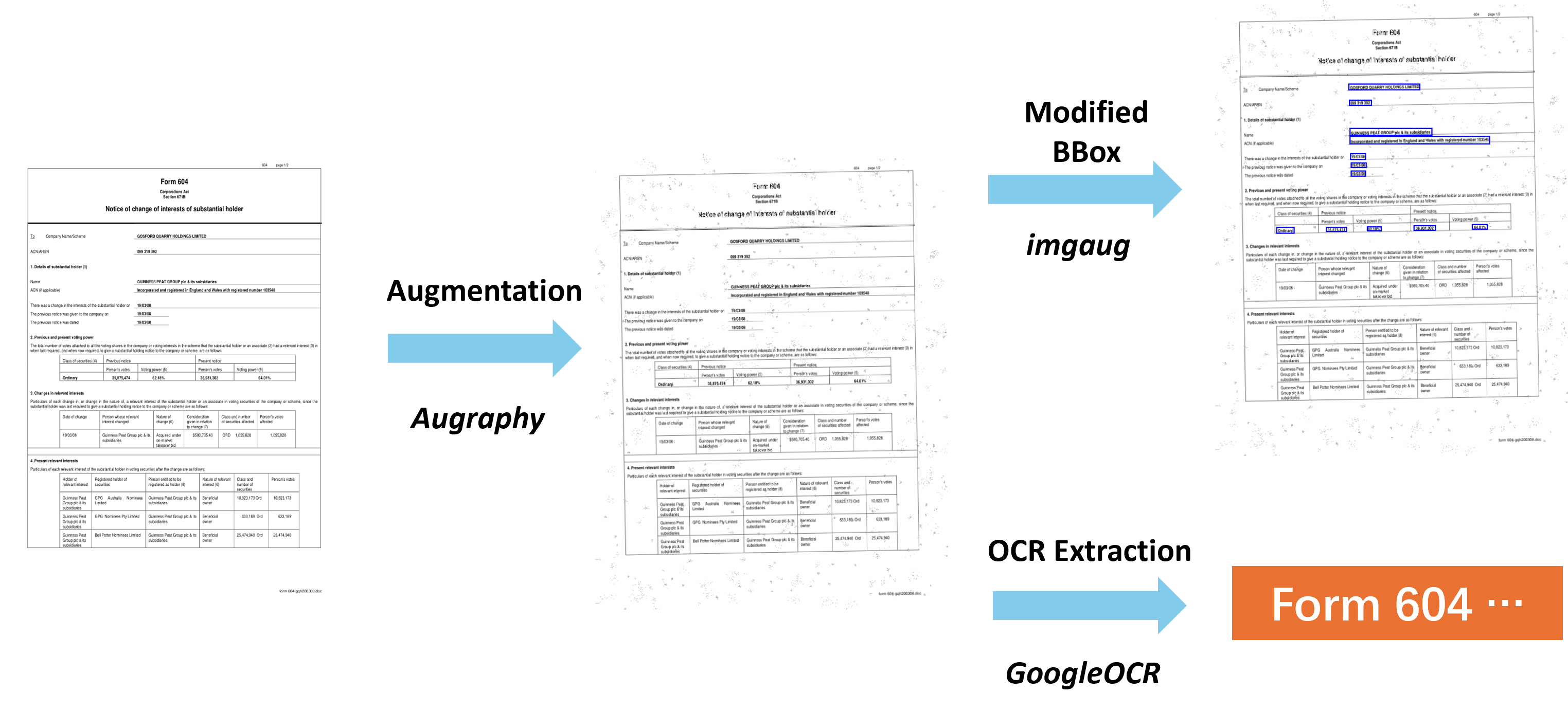}
    \caption{An example of document augmentation from clean digital style to handwritten scan style}
    \label{fig:augment}
\end{figure*}

An overview of the data preprocessing pipeline is shown in Fig. \ref{fig:augment}. We provide a detailed explanation of the method in this section.
\subsubsection{Data Augmentation}
To address the different aspects of digital and handwritten documents, we use Augraphy to generate various document types. However, not all of these augmentations produce documents resembling the scanned handwritten documents in Form-NLU. After careful examination, six augmentations were selected: InkBleed, Letterpress, LowInkRandomLines, LowInkPeriodicLines, JPEG, and DirtyScreen. The first four are text effects, while the last two are paper effects. InkBleed and Letterpress together emulate handwritten characteristics; InkBleed mimics ink bleed, while Letterpress simulates the uneven pressing of ink on paper. LowInkRandomLines and LowInkPeriodicLines introduce low-ink lines to replicate poor print quality. JPEG introduces compression artifacts to simulate the degraded quality of a JPEG file, and DirtyScreen mimics the dirty screen effect often seen in scanned documents. In addition to the effects generated by Augraphy, image rotation is also employed since handwritten documents are typically not scanned perfectly vertically. For each document in the training and validation sets, five augmented counterparts are generated. Each counterpart applies one of the text effects and one of the paper effects, with a 70\% chance of independently applying each effect, along with a 50\% chance of applying a random rotation between -5 to 5 degrees.

\subsubsection{Annotation Preprocessing}
The dataset includes 12 unique keys, each paired with a value of -1 or a bounding box, where -1 indicates the absence of that information in the document. Objects with a value of -1 are removed during training and validation. Since the original bounding boxes would be incorrect after rotation, imgaug \cite{imgaug} is used to obtain the corresponding rotated bounding boxes. Additionally, GoogleOCR\footnote{https://cloud.google.com/vision/docs/ocr} is utilized to extract OCR text for multimodal models that require both image and text inputs.

\subsection{Models}
We tested our pipeline with various backbone models and detection frameworks, including ResNet-101 \cite{he2016deep}, DiT, and LayoutLMv3, using both Faster R-CNN \cite{ren2015faster} and Mask R-CNN \cite{he2017mask} framework.

\textbf{ResNet} is a convolutional neural network that serves as the backbone in the original papers for Faster R-CNN and Mask R-CNN.

\textbf{Dit} is an image transformer model that processes image patches as input, while \textbf{LayoutLMv3} is a multimodal transformer model that takes both text tokens and image patches as inputs. Both DiT and LayoutLMv3 are pre-trained on document images, providing them with domain-specific knowledge, whereas ResNet is pre-trained on natural images, leading to differences in domain expertise.

Faster R-CNN differs from Mask R-CNN in that the latter extends Faster R-CNN by adding a branch for predicting object masks.

\subsection{Evaluation Metrics}
Following VRDIU competion, the evaluation metrics is the mean average percesion (mAP),
\begin{equation}
    mAP_{IoU} = \frac{1}{|C|}\sum _{i=1}^CAP_{IoU}^{(i)}
\end{equation}
where $C$ is set of classes.

\section{Result}
The results of our proposed pipeline are presented in Table \ref{tab:res}. As shown in the table, document augmentation leads to improved information localization performance in handwritten documents for three out of four models,  with a maximum improvement of 3.97\%. Notably, when comparing different backbone models, the performance difference on handwritten documents is significantly larger. For instance, when comparing ResNet and DiT, the performance difference on digital documents is only 4.77\%, while it reaches 27.26\% on handwritten documents. This highlights that a backbone model pre-trained on document images has much stronger generalization ability than the one pre-trained on natural images. Interestingly, LayoutLMv3 achieves higher performance on digital documents but lower performance on handwritten documents, likely due to OCR errors in handwritten text. Overall, these results demonstrate that our augmentation method effectively mimics the domain of handwritten documents using only digital counterparts.

% \usepackage{graphicx}
\begin{table}[]
\small
\begin{tabular}{llllll}
\toprule
\textbf{} & \textbf{} & \multicolumn{2}{c}{\textbf{w/o aug}} & \multicolumn{2}{c}{\textbf{w. aug}} \\ \midrule
\textbf{backbone} & \textbf{D. F.} & \textbf{digital} & \textbf{h.d.} & \textbf{digital} & \textbf{h.d.} \\ \midrule
Resnet-101 & Faster & 0.7239 & 0.2799 & 0.7063 & \textbf{0.3196} \\
Resnet-101 & Mask & 0.7354 & \textbf{0.3566} & 0.7298 & 0.3231 \\
Dit & Mask & 0.7766 & 0.5838 & 0.7775 & \textbf{0.5957} \\
LayoutLMv3 & Mask & 0.7806 & 0.4722 & 0.7781 & \textbf{0.5092}\\\bottomrule
\end{tabular}%
\caption{Generalization Performance w/wo Data Augmentation\\ (D. F.: Detection Frame, Faster and Mask stands for Faster R-CNN and Mask R-CNN; h.d.: handwritten; w/o: without, w.: with)}
\label{tab:res}
\end{table}
\section{Conclusion}
This study addresses the challenge presented by VRDIU Track B, which focuses on localizing key information in document images. Our pipeline includes a data augmentation phase that simulates handwritten documents from digital ones, followed by an object detection phase. Through experiments, we demonstrate that the data augmentation technique can enhance models' generalization performance on unseen handwritten datasets. In the future, we plan to integrate unsupervised learning methods to enable the model to acquire different domain knowledge before undertaking cross-domain tasks.
\section{Implementation Details}
All experiments were conducted using 2 RTX 4090 (24GB) GPUs. Each model was trained for 10,000 steps with a learning rate of 5.0e-05, and the best model was selected based on the mAP on the validation set.

\bibliography{aaai25}

\end{document}